\documentclass[10pt,twocolumn,letterpaper]{article}

\usepackage{wacv}
\usepackage{times}
\usepackage{epsfig}
\usepackage{graphicx}
\usepackage{amsmath}
\usepackage{amssymb}
\usepackage{bm} 
\usepackage{hyperref}
\usepackage{float}
\hypersetup{
    colorlinks=true,
    linkcolor=blue,
    filecolor=magenta,      
    urlcolor=cyan,
}

\wacvfinalcopy 

\def\wacvPaperID{739} 

\ifwacvfinal\fi
\begin{document}

\title{Relighting Images in the Wild with a Self-Supervised Siamese Auto-Encoder}

\author{Yang Liu \\
 University of Surrey, UK\\
{\tt\small yang.liu@surrey.ac.uk}
\and
Alexandros Neophytou, Sunando Sengupta, Eric Sommerlade \\
Microsoft Corporation, Reading, UK\\
{\tt\small Alexandros.Neophytou, Sunando.Sengupta }\\
{\tt\small  Eric.Sommerlade@microsoft.com}
}

\maketitle

\begin{abstract}
We propose a self-supervised method for image relighting of single view images in the wild. The method is based on an auto-encoder which deconstructs an image into two separate encodings, relating to the scene illumination and content, respectively. In order to disentangle this embedding information without supervision, we exploit the assumption that some augmentation operations do not affect the image content and only affect the direction of the light. A novel loss function, called spherical harmonic loss, is introduced that forces the illumination embedding to convert to a spherical harmonic vector. We train our model on large-scale datasets such as Youtube 8M and CelebA. Our experiments show that our method can correctly estimate scene illumination and realistically re-light input images, without any supervision or a prior shape model. Compared to supervised methods, our approach has similar performance and avoids common lighting artifacts.
\end{abstract}

\section{Introduction}

Relighting images in the wild has gained popularity recently, especially since the development of mobile computing and video communication has led to an explosion in the consumption of digital photography. The diversity of the application environments, e.g. indoor, outdoor, day or night, makes the task of realistically relighting images challenging. In an ideal use case scenario, users can choose the desired illumination of an image, without having to consider the illumination of the original image. However, even state-of-art lighting algorithms meet three main problems. The first problem is the lack of large-scale relighting datasets since it is hard to manually label scene illumination, especially when there is more than one light source. The collection of image annotations has been the main bottleneck of many supervised relighting methods. The second problem is that most relighting algorithms need multiple views of the same object for training, which hinders the algorithms from learning from wild data. The third problem is that relighting usually requires depth information to avoid artifacts from shadows or over-lighting.

\begin{figure}[htbp]
    \centering
    \includegraphics[width=8 cm]{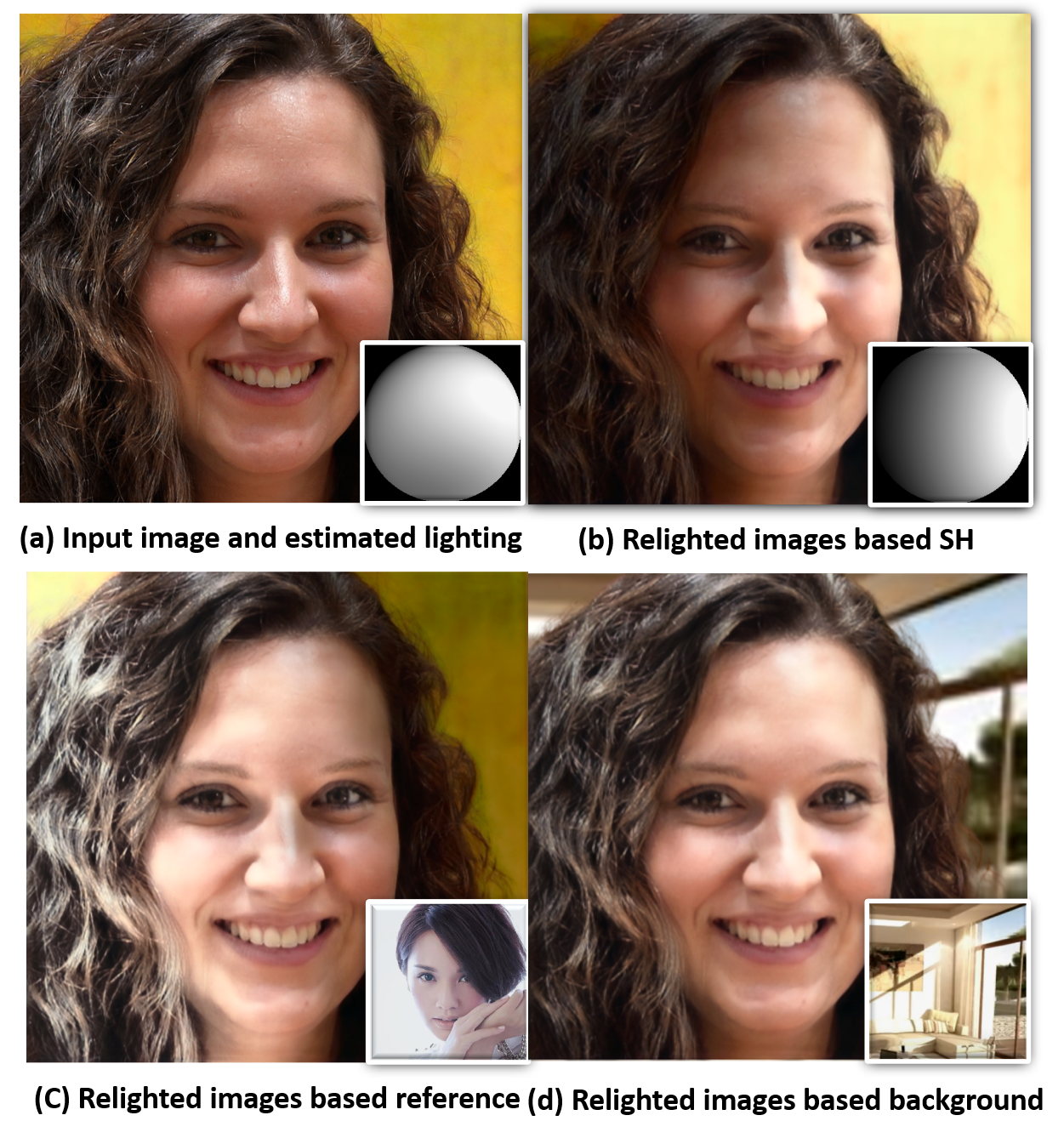}
    \caption{Relighted images with different target illumination conditions. Our method can relight an image (a) based on given light direction via  spherical harmonic coefficients (b) as well as estimated illumination from a reference face (c) and environment scenes (d). }
    \label{fig:fig1}
\end{figure}
The goal of this work is to design a single automatic image relighting network with a large-scale single-view unlabelled dataset. It takes a single image and target lighting as inputs then estimates the lighting of the input image and subsequently generates a new, relighted image based on the target lighting. Specifically, our approach uses a self-supervised auto-encoder network which decomposes the image into two embeddings: one for illumination and one for content. There are two main challenges to this idea. The first challenge is how to separate illumination information from content information correctly. To address this, we can augment images in such a way that the geometry of the objects stays the same while the direction of the light changes. To satisfy this assumption, we consider four possible augmentations: two flipped images, a rotated image and an inverted image. Each augmentation image is paired up with the original image and used to train a Siamese auto-encoder network. We assume that the image training pairs have the same content but different illumination. Based on this, we can decouple the image content embedding from the image illumination embedding. The second challenge is how to generate a semantically meaningful light representation.  Without ground truth light information, the illumination embedding has a large number of possibilities. It is impossible to relight images by adjusting the embedding manually. Our solution is to design a spherical harmonic (SH) loss that forces the illumination embedding to take the form of Laplace's spherical harmonics. When Spherical Harmonics \cite{macrobert1947spherical} represent the illumination embedding, the relighting can be meaningfully controlled. 

The contribution of our work is two-fold: First, we propose a relighting self-supervised auto-encoder network, which is trained on image pairs and separates images into content and lighting embeddings. Without ground-truth illumination information, the proposed method can generate high-resolution (1024x1024) relighted images.
Second, a novel spherical harmonics loss is introduced, which is based on the assumption that the proposed image augmentations only affect the direction of the light in the images and not the geometry of the objects. The relighting can be controlled by adjusting the values in illumination embedding. Finally, we test our method on both synthetic and real data. We show that we achieve higher fidelity in the relighted images compared to other supervised learning and self-supervised methods.

\section{Related Work}

\subsection{Image Relighting}

Relighting, especially for synthetic objects, has been a popular task for over twenty years \cite{liu2020learning, yu2020self}. Most of the image relighting approaches which were introduced addressed face portrait images. With relative geometry information, landmarks and detectors of human faces, Zhou et al. \cite{zhou2019deep} accurately estimate surface normals required for relighting from a single image based on an SFSNet network \cite{sengupta2018sfsnet}. Sun et al. \cite{sun2019single} use a similar architecture, but the encoder extracts lighting features in addition to facial features. However, the objective functions are only applied to masked people and the background are simply replaced by blurred environment maps. While much of the previous works focus on portrait relighting, only a few papers explicitly consider illumination estimation from environment images. Duchene et al. \cite{duchene2015multi} propose an outdoor scene relighting network, which removes the shadows of the input images and adds new shadows for the relighted images based on a Markov Random Field over a graph of points. Philip et al. \cite{philip2019multi} further extend this idea by casting shadows from a 3D geometric prior. A SLAM algorithm is applied to estimate the position of the main light sources based on the scene’s geometry and specular regions \cite{whelan2016elasticfusion}. Zhang et al. \cite{zhang2016emptying} use RGB-D data to relight the indoor environment and estimates the materials of the furniture. However, most of the previous works need 2D/3D geometric priors or RGB-D sensors. Relighting a single-view scene image is still a challenging problem.

\subsection{Photo Style Transfer}

Photo style transfer, where the input is a source image and a reference image and the output is the illumination inversion image with the style of the reference image \cite{shih2014style, shu2017portrait}, is similar to image relighting. The most challenging problem of a machine learning method is the difficulty in designing a suitable loss function. Since generative adversarial networks (GANs) \cite{goodfellow2014generative} can estimate the loss from the training data, GANs are widely used in different photo style translation problems, such as pix2pix \cite{isola2017image} and CycleGAN \cite{zhu2017unpaired}. Instead of training the model with loss function, the generator and discriminator are trained. The generator learns to produce data similar to the training data, and discriminator learns to find the mistake of the generator. Auto-encoder is also applied in the photo style transfer \cite{shu2017neural, li2019learning, dherse2020scene}. For example, Shu et al. \cite{shu2017neural} propose a physically grounded rendering-based disentangling network specifically designed for faces. Landmarks and 3D face model are applied with an auto-encoder for estimation of face illumination. Li et al. \cite{dherse2020scene} proposes a linear propagation module with a transformation to enable a feed-forward network for photo-realistic style transfer. However, the object shape in the transformed images is fuzzy, and the reference image affects the colour of transformed images.



\subsection{Siamese network}

The Siamese network architecture was proposed in the 1990s to solve signature verification as an image matching problem \cite{bromley1994signature}, single-target tracking \cite{li2019siamrpn++, bertinetto2016fully} and one-shot learning \cite{koch2015siamese}. The pre-trained Siamese CNN features have been used with Context-RNNGAN model for image generation \cite{choi2018structured}. Compared to the above work, the relighting auto-encoder would replace the CNN network in the Siamese network, which is trained without external supervision. To the extent of our knowledge, our work is the first  using a Siamese network for relighting.

\section{The relighting network}

Our goal is to learn an image relighting model from large unlabelled image datasets. The model receives a source image and target illumination as input and outputs an estimation of the source image illumination and the relighted image. In this section, we provide the implementation details for the self-supervised relighting auto-encoder. Since the ground truth illumination information and relighted images are not available for wild datasets, the objective function and training detail of the relighting auto-encoder is further discussed. 

\subsection{Relighting Auto-encoding}

\begin{figure}
    \centering
    \includegraphics[width=8 cm]{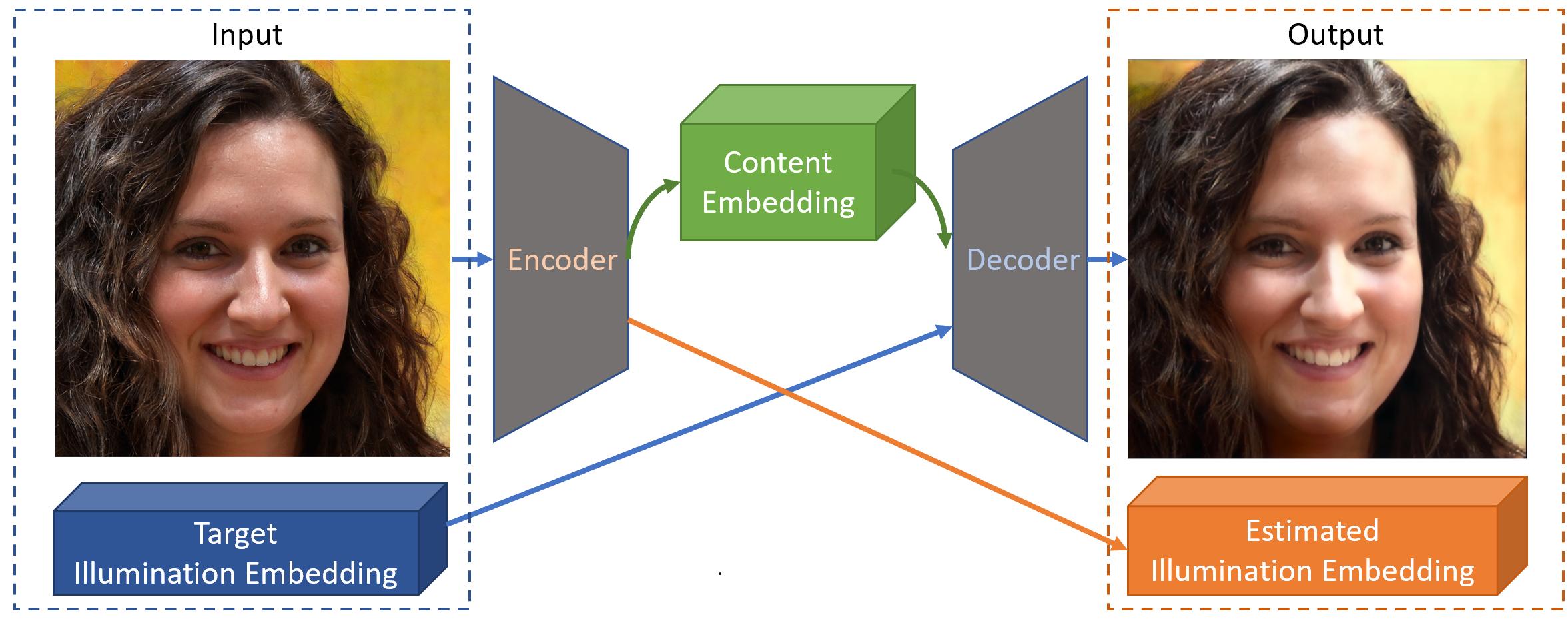}
    \caption{Auto-encoder: The input is the original image and target illumination embedding. The output is the re-lit image with the target illumination and estimated illumination embedding of the input image. }
    \label{fig:ae block}
\end{figure}

The relighting network is an auto-encoder (AE), as shown in Fig. \ref{fig:ae block}. With the encoder $\bm{E}$, the input image $\bm{I}_{\bm{L}}$ is decomposed to a content embedding $\bm{C}$ (green boxes) and an illumination embedding $\hat{\bm{L}}$ (orange boxes). $\bm{I}_{\bm{L}} \in \mathbb{R}^{w \times h \times 3}$ is the image $\bm{I}$ with illumination embedding $\bm{L}$, where $w$ and $h$ are the image weight and height. $\bm{C} \in \mathbb{R} ^{m \times m \times d} $ and $\hat{\bm{L}}\in \mathbb{R}^{n}$ are two tensors, where $n$ is the size of the illumination embedding, $m$ is the size of the content embedding and $d$ is the depth of the content embedding. The encoder network can be represented as:
\begin{equation}
\{ \bm{C}, \hat{\bm{L}}\} = \bm{E}(\bm{I}_{\bm{L}}) 
\end{equation}
With the decoder $\bm{D}$, the content embedding $\bm{C}$ and target illumination embedding $\bm{L}'$ (blue boxes) are used to rebuild the relighted image $\hat{\bm{I}}_{{\bm{L}}'}$. The decoder network can be represented as:
\begin{equation}
\hat{\bm{I}}_{{\bm{L}}'} = \bm{D}(\bm{C}, {\bm{L}}')
\end{equation}

If the target and source illumination embedding are the same, the AE network becomes a reconstruction network. Otherwise, it is a relighting network. For training the relighting auto-encoder, we employ three objective functions. These are reconstruction loss, Spherical Harmonic loss and discriminator loss. 

\subsection{Reconstruction Loss}

\begin{figure}
    \centering
    \includegraphics[width=8 cm]{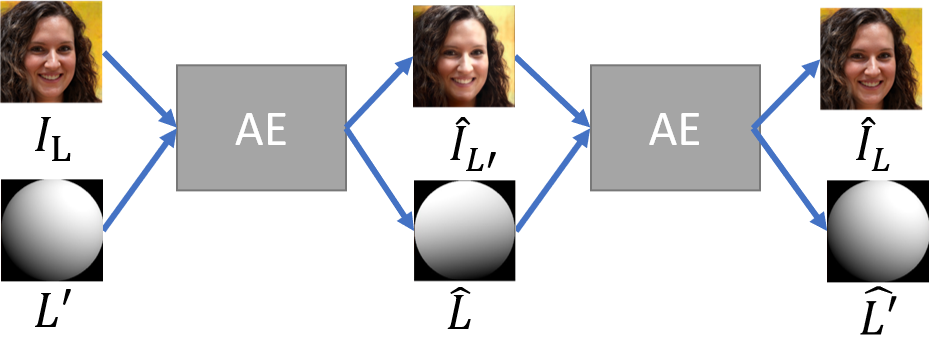}
    \caption{The Siamese reconstruction network. Two auto-encoders have the same structure, and they share weights. The input is the source image $\bm{I}_{\bm{L}}$ while the output is the reconstructed image $\hat{\bm{I}}_{\bm{L}}$. $\bm{L}'$ is a random lighting illumination while $\hat{\bm{L}'}$ is the estimated illumination. } 
    \label{fig:R loss}
\end{figure}

An auto-encoder network can learn $\bm{E}$ and $\bm{D}$  that can relight images from a large number of relighted image pairs. However, the ground truth illumination embedding and relighted images are not available for the wild data. To address this problem, two AE networks form a reconstruction network, as shown in Fig. \ref{fig:R loss}. The two auto-encoders have the same structure and their weights are shared.  Therefore, these two auto-encoders are called as Siamese auto-encoder. $\bm{I}_{\bm{L}}$ is an image in the training dataset. $\bm{L}'$ is a random illumination embedding. $\hat{\bm{I}}_{\bm{L}'}$ and $\hat{\bm{I}}_{\bm{L}}$ is the transferred image with the illumination $\bm{L}'$ and the estimated illumination $\hat{\bm{L}}$, respectively. $\hat{\bm{L}}$ and $\hat{\bm{L}'}$ is the estimated illumination from $\bm{I}_{\bm{L}}$ and $\hat{\bm{I}}_{\bm{L}'}$, respectively. With the two transformation networks, the reconstruction network is setup. The network then takes source image $\bm{I}_{\bm{L}}$ and target lighting $\bm{L}'$ as input and generates $\hat{\bm{I}}_{\bm{L}}$ and $\hat{\bm{L}'}$. $\bm{I}_{\bm{L}}$ and $\bm{L}'$ are used as ground truth to supervise the training in this reconstruction task. We use mean absolute error loss for the reconstructed image $\hat{\bm{I}}_{\bm{L}}$ and the estimated lighting embedding $\hat{\bm{L}'}$. The image gradient is also considered with mean absolute error loss to preserve edges and avoid blurring. The objective for this reconstruction network is,
\begin{equation}
\begin{aligned}
      \mathcal{L}_{roc}(\bm{I}_{\bm{L}}, \bm{L}') = &\frac{1}{w*h}(\left \| \bm{I}_{\bm{L}} - \hat{\bm{I}}_{\bm{L}} \right \|_1 + \left \| \nabla\bm{I}_{\bm{L}} - \nabla\hat{\bm{I}}_{\bm{L}} \right \|_1) 
    \\ &+ \frac{1}{n}\left \| \bm{L}' - \hat{\bm{L}'}\right \|_1  
\end{aligned}
\end{equation}
$\mathcal{L}_{roc}(\bm{I}_{\bm{L}}, \bm{L}')$ is simply represented as $\mathcal{L}_{roc}(\bm{I}_{\bm{L}})$, since $\bm{L}'$ is given and fixed in each epoch.


\subsection{Spherical Harmonic Loss}
Although the reconstruction loss can let the AE network converge, $\hat{\bm{I}}_{\bm{L}'}$ and $\hat{\bm{L}}$ lack constraints. The AE network can map the same image and illumination embedding to any random image and illumination in the target domain. Any of the learned mappings can induce an output distribution that matches the target distribution. Thus, reconstruction losses alone cannot guarantee that the AE network can relight images. To further reduce the space of possible mapping functions, some constraints should be applied for $\hat{\bm{I}}_{\bm{L}'}$ and $\hat{\bm{L}}$. A  parametrization widely used in relighting tasks is a Spherical Harmonic, defined on the surface of a sphere. The details about calculating 
Spherical Harmonic can be found in \cite{macrobert1947spherical}. Here we introduce a novel loss function, based on Spherical Harmonics, as a constraint on $\hat{\bm{I}}_{\bm{L}'}$.

\begin{figure*}
    \centering
    \includegraphics[width=16cm]{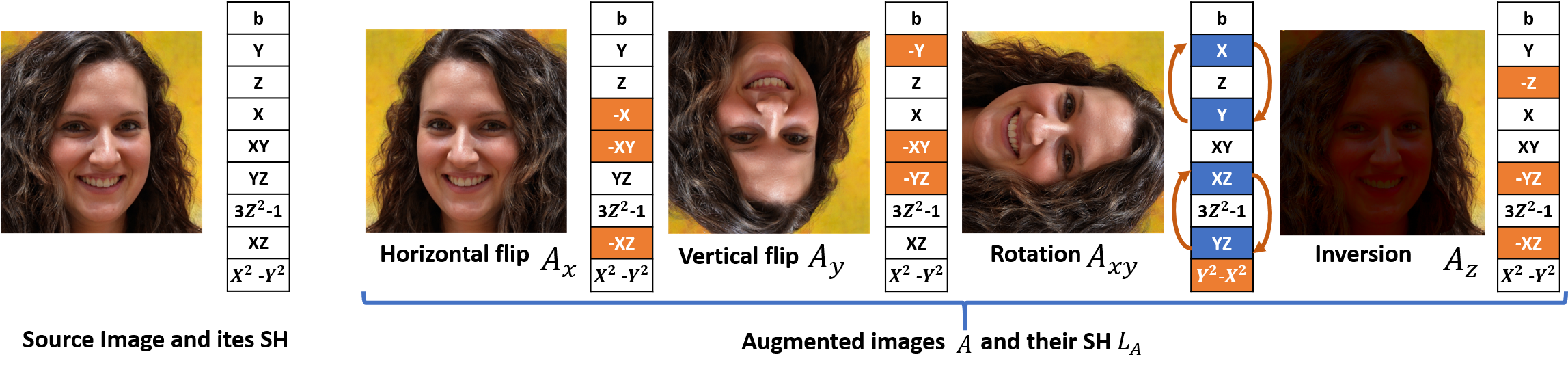}
    \caption{Augmentation images: the original image is shown at the bottom. The definition of Spherical Harmonic illumination embedding is given above the original image. The four kinds of augmented images are shown $\bm{A}_x$, $\bm{A}_y$, $\bm{A}_{xy}$ and $\bm{A}_z$, which are obtained as horizontal flip, vertical flip, rotation and inversion. These augmentations change some channels of the Spherical Harmonic illumination embedding of image. The changed illumination embedding is given below the augmented images, where the white channel is unchanged. The value in the orange channel turns negative. The blue channel is swapped with other blue channels as the orange lines.}  
    \label{fig:augemented}
\end{figure*}

Spherical Harmonic loss is a novel loss to guarantee that the illumination embedding is represented by Spherical Harmonic lighting. For controlling the illumination, Spherical Harmonic lighting is used to represent the illumination, shown above the original image in Fig. \ref{fig:augemented}, where $b$ is the bias and its value is $\frac{\sqrt{\pi}}{2}$. $X$, $Y$, $Z$ means the channels is linearly dependent on the $x$, $y$ and $z$ axis in the space. Since the Spherical Harmonics of the training data is unknown, some augmented images $\bm{A}$ are used as the associated images to calculate the Spherical Harmonic Loss. Four image augmentations are considered. A horizontally flipped image $\bm{A}_x$, a vertically flipped image $\bm{A}_y$, a rotated 90 degrees image $\bm{A}_{xy}$, and an inverted image $\bm{A}_z$ (discussed later), as shown in Fig. \ref{fig:augemented}. These augmentations change some values in the illumination embedding. The changes are given below each augmented images. The unchanged channel is shown as the white block, while the value in the orange block turns negative. Since $\bm{A}_{xy}$ is given by rotation, the order of some channels (blue channels) is changed. Specifically, the 2-nd and 6-th channels are swapped with 4-th and 8-th channels, respectively. Although $\bm{A}_y$ and $\bm{A}_xy$ introduces unnatural upside down or tilted images, this does not make the job of the auto-encoder more difficult in our experiments.  
\begin{figure}
    \centering
    \includegraphics[width=8cm]{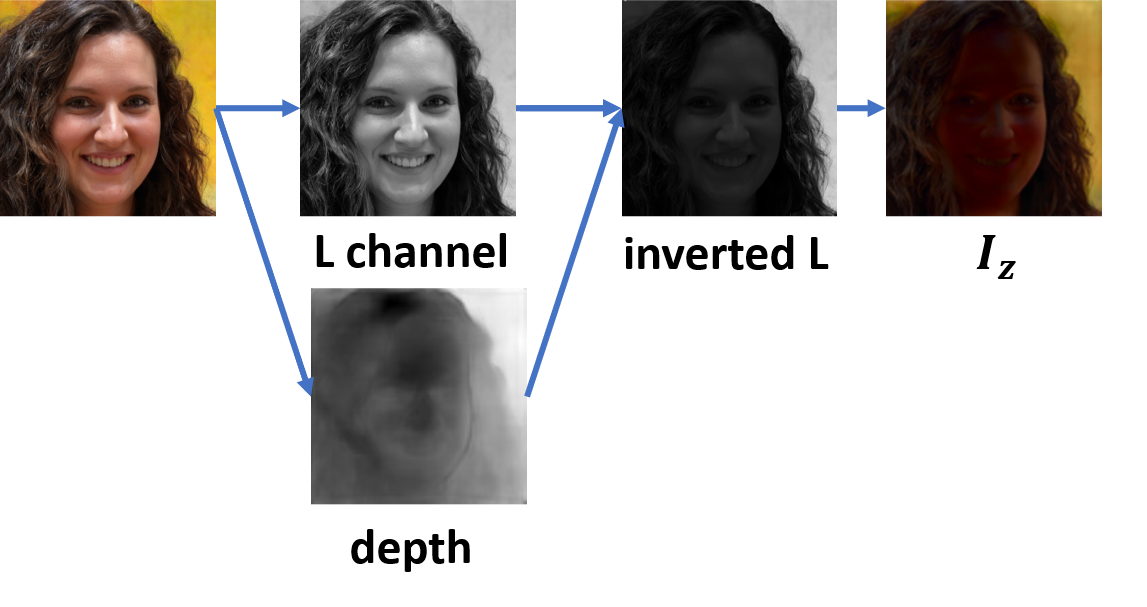}
    \caption{$\bm{A}_z$ pipeline: the "L channel" and "depth" information are estimated from the input LAB image. The "L channel" and  "depth" information are further merged into "inverted L" as Eq. \eqref{eq:ml}. Finally, $\bm{A}_z$ is build up from the "inverted L" and the AB channel of input image.} 
    \label{fig:Iz}
\end{figure}

Different from $\bm{A}_x$, $\bm{A}_y$ and $\bm{A}_{xy}$ which can be calculated based on the flip and rotation operations, the inversion augmentation $\bm{A}_{z}$ is using a more sophisticated approach, as we need to approximate a lighting change in depth direction. We include a pretrained intrinsic image network, similar to SFSnet \cite{sengupta2018sfsnet}, since the illumination information in depth can not be directly given in the input images. The process of estimation of $\bm{A}_z$ is shown in Fig. \ref{fig:Iz}. Firstly, the input image is converted from RGB into CIELAB (LAB) colour space. The L channel of the input image is shown as "L channel". Secondly, the depth information, shown as "depth", is estimated from the input image by a pre-trained depth prediction model \cite{chen2016single}. The depth value of $\bm{I}$ is shown as $\mathcal{D}(\bm{I})$. The depth information is applied to separate the foreground objects (faces, cars and so on) from the background. The pixels with low depth value belong to the objects while pixels with high depth value belong to the background. Apart from that, we assume light sources are normally located at the front of the objects. When the light sources are moved to the back of the objects, the luminance of objects decreases while the luminance of background stay the same since it has different light sources. The L channel of input images should be inverted based on the depth information to calculate $\bm{A}_z$. We tested a multitude of inversion approximations and found the following inversion equation to visually provide the best results:
\begin{equation}
\label{eq:ml}
    \bm{\mathbb{L}}(\bm{A}_z) = \frac{\left \| (\bm{1}_{w*h}-\mathcal{D}(\bm{I}))\circ \bm{\mathbb{L}}(\bm{I}) +
    \mathcal{D}(\bm{I}) \circ \text{tanh}(\bm{\mathbb{L}}(\bm{I}))/2  \right \|_1}{w*h}
\end{equation}
where $\circ$ is the Hadamard product. $\bm{\mathbb{L}}(.)$ is function to calculate the L channel information in LAB colour space and $\text{tanh}$ is a tanh function. $\bm{A}_z$ is the image with $\bm{\mathbb{L}}(\bm{A}_z)$. $\tanh(.)/2$ is the scale function, which can be adjusted for different tasks.

\begin{figure}
    \centering
    \includegraphics[width=5 cm]{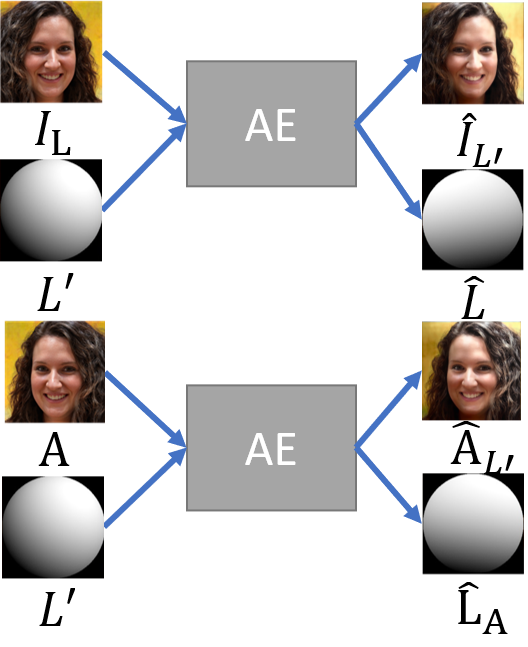}
    \caption{The Siamese comparison network. The two auto-encoder have the same structure, and they share weights. The input is source image $\bm{I}_{\bm{L}}$ and augmented image $\bm{A}$ while the output is the illumination estimation $\hat{\bm{L}}$ and $\hat{\bm{L}}_{\bm{A}}.$} 
    \label{fig:A loss}
\end{figure}

Based on the above properties, the relighting task is converted to a comparison task, as shown in Fig. \ref{fig:A loss}. $\bm{I}_{\bm{L}}$ is a random image in the dataset, whose illumination embedding is the $\bm{L}$ and $\bm{L}'$ is the random illumination embedding. The augmented image of ${\bm{I}}_{\bm{L}}$ is shown as the $\bm{A}$, where $\bm{A} \in \{ \bm{A}_x, \bm{A}_y, \bm{A}_{xy}, \bm{A}_{z}\}$. The relighted ${\bm{I}}_{\bm{L}}$ and $\bm{A}$ with the target illumination embedding $\bm{L}'$ are shown as $\hat{\bm{I}}_{\bm{L}'}$ and $\hat{\bm{A}}_{\bm{L}'}$, respectively. The estimated illumination embedding of ${\bm{I}}_{\bm{L}}$ and $\bm{A}$ are $\hat{\bm{L}}$ and $\hat{\bm{L}}_{\bm{A}}$. By comparing $\hat{\bm{L}}$ and $\hat{\bm{L}}_{\bm{A}}$, Spherical Harmonic loss $\mathcal{L}_{sh}(\bm{I}_{\bm{L}}, \bm{A})$ is defined as:
\begin{equation}
\label{eq:shloss}
\begin{aligned}
\mathcal{L}_{sh}(\bm{I}_{\bm{L}}, \bm{A}) = \bm{1}^T_{9*1}\hat{\bm{L}} + \mathcal{C}(\bm{A})\hat{\bm{L}}_{\bm{A}} -  \sqrt{\pi}
\end{aligned}
\end{equation}
where
\begin{equation}
\mathcal{C}(\bm{A})=
\begin{cases} 
[1, -1, -1, 1, 1, -1, -1, 1, -1]^T & \bm{A} = \bm{A}_x \\
[1,  1, -1,-1, 1,  1, -1,-1, -1]^T & \bm{A} = \bm{A}_y \\
[1, -1, -1,-1, -1, -1, -1,-1,  1]^T & \bm{A} = \bm{A}_{xy} \\
[1, -1,  1,-1,-1,  1, -1, 1, -1]^T & \bm{A} = \bm{A}_z
\end{cases}
\end{equation}
where $\bm{1}^T_{9 \times 1}$ is a $9 \times 1$ vectors, where each element value is 1. Eq. \eqref{eq:shloss} is introduced based on the relationship between the input image $\bm{I}_{\bm{L}}$ and its augmented image $\bm{A}$, as shown Fig. \ref{fig:augemented}. The sum of first channel of $\bm{I}_{\bm{L}}$ and $\bm{A}$ should be equal to $\sqrt{\pi}$, since the bias is $\sqrt{\pi}/2$. For 2-9 channel of the illumination embedding of $\bm{I}_{\bm{L}}$ and $\bm{A}$, the different of the unchanged channel and sum of the unchanged channel are both equal to 0.

\subsection{Discriminator loss}

Since our network is trained using image flipping and rotation, they may contain artifacts due to inaccurate estimation of object depth and lighting. Apart from that, the relighted images lack the constraint condition. A GAN loss is proposed to improve the quality of the generated images. WGAN-GP is applied to force the distribution of local image
patches to be close to that of a natural image. The objective is given as:

\begin{equation}
    \mathcal{L}_{dis}(\bm{I}_{\bm{L}}) = \mathbb{E}_{\bm{I}_{\bm{L}}} \mathcal{C}\left(\bm{D}(\bm{E}(\bm{I}_{\bm{L}}), \bm{L}')\right)^{2} -\mathbb{E}_{\bm{I}_{\bm{L}}}(\mathcal{C}(\bm{I}_{\bm{L}}))^{2}
\end{equation}
where $\mathcal{C}$ is the critic (discriminator) where the re-light images are $\bm{E}(\bm{I}_{\bm{L}})$ and ${\bm{I}}_{\bm{L}}$ are the real images. 

\subsection{Implementation Details}

The overall loss for our network is a linear combination of the losses mentioned:
\begin{equation}
\begin{aligned}
    \mathcal{L} = &\mathcal{L}_{roc}(\bm{I}_{\bm{L}}) + \alpha \mathcal{L}_{dis}(\bm{I}_{\bm{L}})\\
    &+\frac{1}{4} \sum_{\bm{A}\in \{\bm{A}_x, \bm{A}_y, \bm{A}_{xy, \bm{A}_z}\}}\mathcal{L}_{roc}(\bm{A}) + \beta \mathcal{L}_{sh}(\bm{I}_{\bm{L}}, \bm{A})
\end{aligned}
\end{equation}
where $\alpha = 0.5$ and $\beta = 0.25$. We train our network with images of resolution 1024 $\times$ 1024. More specifically, the source images pass through six down-sampling layers and eight residual blocks. Then the embedding passes through three residual blocks and two residual blocks with a fully-connected layer to get the content and illumination embeddings, respectively. The size $m$ and depth $d$ of the content embedding are set as 64 and 512, respectively. Then these embeddings are added after several residual blocks. Finally, a relighted image is generated after six up-sampling layers. Since the encoder losses some information present in the input images, the reconstructed image appears blurry. Therefore, six skip layers are added between the down-sampling and up-sampling layers. We train our network for 5000 epochs (about 300 hours) using the Adam optimizer with default parameters.

\section{Experiments}

In this section, we will evaluate our proposed method performance and compare it with previous state-of-the-art methods. Since our network can predict lighting, the model is evaluated in two ways: (A) Given a source image and an SH lighting, relighting an image (denoted as the SH-based relighting). (B) Given a source image and a reference image, estimating SH lighting from the reference image and using the estimated lighting to relight the source image (denoted as the image-based relighting). 

\subsection{Setup}

\textbf{Datasets}: We show the effectiveness of the proposed method on three different datasets:  CelebA \cite{liu2018large}, Youtube 8M \cite{abu2016youtube} and synthetic face dataset \cite{kowalski2020config}. CelebA is a large-scale face attributes dataset with about 200k portrait images. The persons shown in the dataset have large pose variation with different backgrounds. CelebA has large diversities and large quantities. However, the ground truth of illumination is not provided. Since image data of CelebA is limited, and we hope the model can accurately estimate the illumination of background. We pre-trained our model on Youtube 8M, where 6 million videos are included. We only choose a subset of the full dataset. The included categories are "bus", "metro", "park", "shopping mall", "street", "road", "office" and "room" which provide common use case scenarios. For each video, a frame is captured for every 1 minute. Finally, there are nearly 1 million scene images. Synthetic face dataset is generated by a synthetic face framework \cite{kowalski2020config}. The illumination can be manually controlled. Therefore, this data is used to evaluate our proposed method. Note that this dataset is not used to train our methods. 

\textbf{Evaluation metric}: Three error metrics are used to measure the relighting performance across our validation set and testing. They are the Root Mean Square Error (RMSE), Structural dissimilarity (DSSIM) \cite{wang2004image}, and scale-invariant RMSE (RMSE-s) \cite{barron2014shape}. RMSE is a common metric for evaluating the reconstruction task. 
DSSIM is invariant to local and global scaling and tinting. The DSSIM implementation uses a 11 $\times$ 11 Gaussian filter with $\alpha$ = 1.5, $k_1 = 0.01$ and $k_2 = 0.03$, as set in \cite{wang2004image}. DSSIM is computed on each RGB channel of the input image and relighted image individually. It is not sensitive to global and local scaling or colour shifts.

RMSE-s is introduced to solve the lighting scale problem (SH is affected by scale factors, such as the exposure time, except the lighting conditions). The RMSE-s solves for the single scale-factor applied to the predicted image,
\begin{equation}
    \operatorname{RMSE-s}( \bm{I}_{\bm{L}}, \hat{\bm{I}}_{\bm{L}})= \frac{1}{w * h}\min _{\alpha}\left\|\bm{I}_{\bm{L}} - \alpha \hat{\bm{I}}_{\bm{L}}\right\|_{2}
\end{equation}
where $\alpha$ is a scalar. This metric solves for a single global scaling rather than a per-channel scaling. It is still sensitive to erroneous tints in the relighting image. 

\subsection{Ablation Study}

\begin{table}
\begin{center}
\begin{tabular}{|c|c|c|c|}
\hline
Model & RMSE ({\tiny{$10^{-3}$}})& DSSIM ({\tiny{$10^{-3}$}}) &RMSE-s ({\tiny{$10^{-3}$}})\\
\hline\hline
(1)Ours full &  $\bm{6.8}$ & $\bm{3.8}$ & $\bm{9.9}$\\
(2)w/o $\mathcal{L}_{roc}$ &  16.2 & 3.9 & 10.0\\
(3)w/o $\mathcal{L}_{sh}$ &  20.3 & 4.3 & 15.8\\
(4)w/o $\bm{A}_x$ &  18.6 & 4.5 & 15.7\\
(5)w/o $\bm{A}_y$ &  19.8 & 3.8 & 17.9\\
(6)w/o $\bm{A}_{xy}$ &  10.7 & 3.8 & 10.8\\
(7)w/o $\bm{A}_{z}$ &  7.3 & 3.8 & 13.1\\
(8)w/o $\mathcal{L}_{dis}$ &  7.1 & 3.9 & 10.4\\
(9)w/o skip &  7.9 & 3.9 & 10.6\\
\hline
\end{tabular}
\end{center}
\caption{Ablation study. Ablated model, which some part of the proposed model is removed, is evaluated.}
\label{tab:ablation}
\end{table}

\begin{figure}[hbtp]
    \centering
c    \includegraphics[width= 7cm]{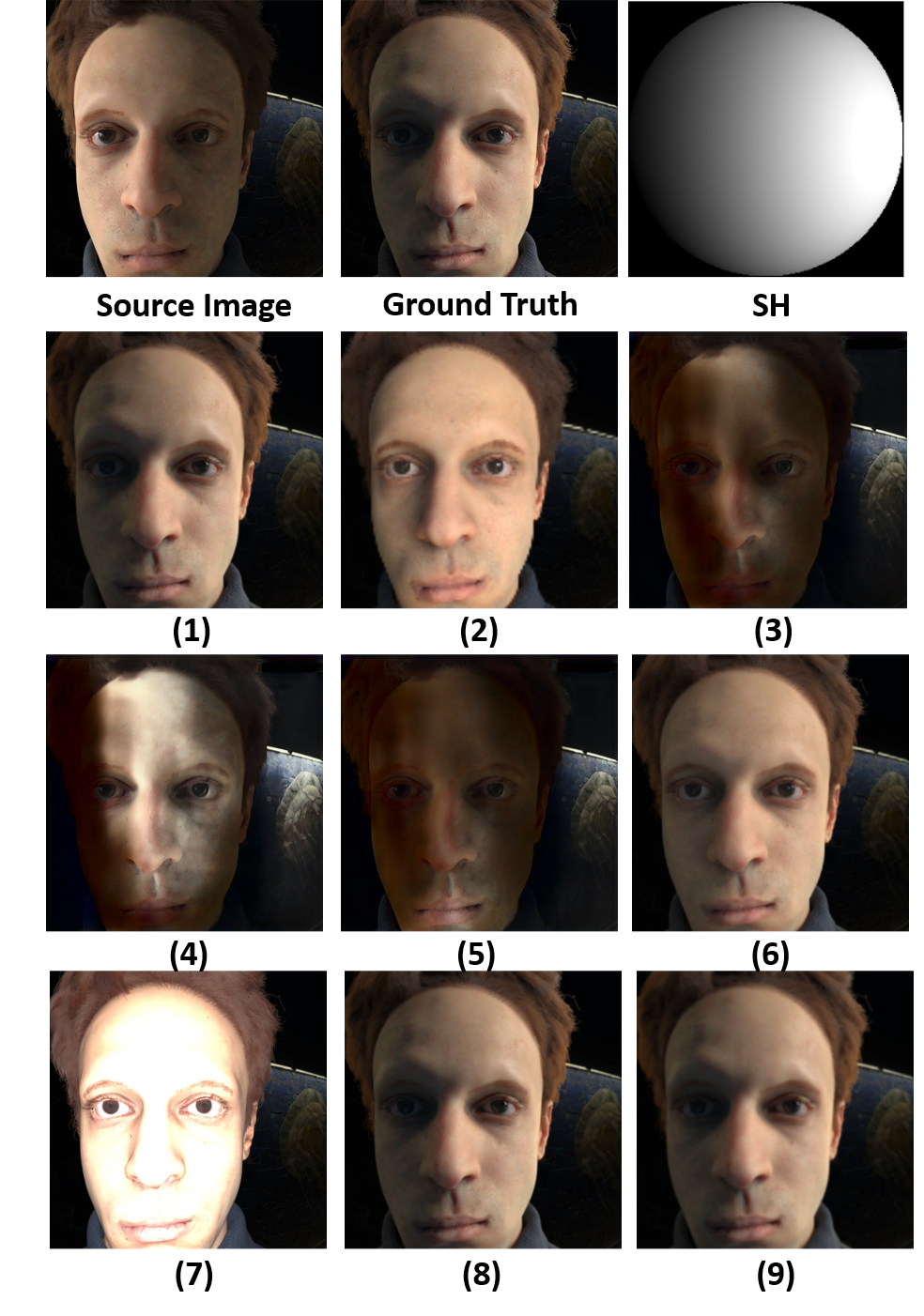}
    \caption{The first row shows the input image, the ground truth of the relighting image and the target SH. (1-9) show the relighted images of our full model, w/o $\mathcal{L}_{roc}$, w/o $\mathcal{L}_{sh}$, w/o $\bm{A}_x$, w/o $\bm{A}_y$, w/o $\bm{A}_{xy}$, w/o $\bm{A}_{z}$, w/o $\mathcal{L}_{dis}$ and w/o skip connection models, respectively.} 
    \label{fig:ablation}
\end{figure}

\begin{figure}[hbtp]
    \centering
    \includegraphics[width=7 cm]{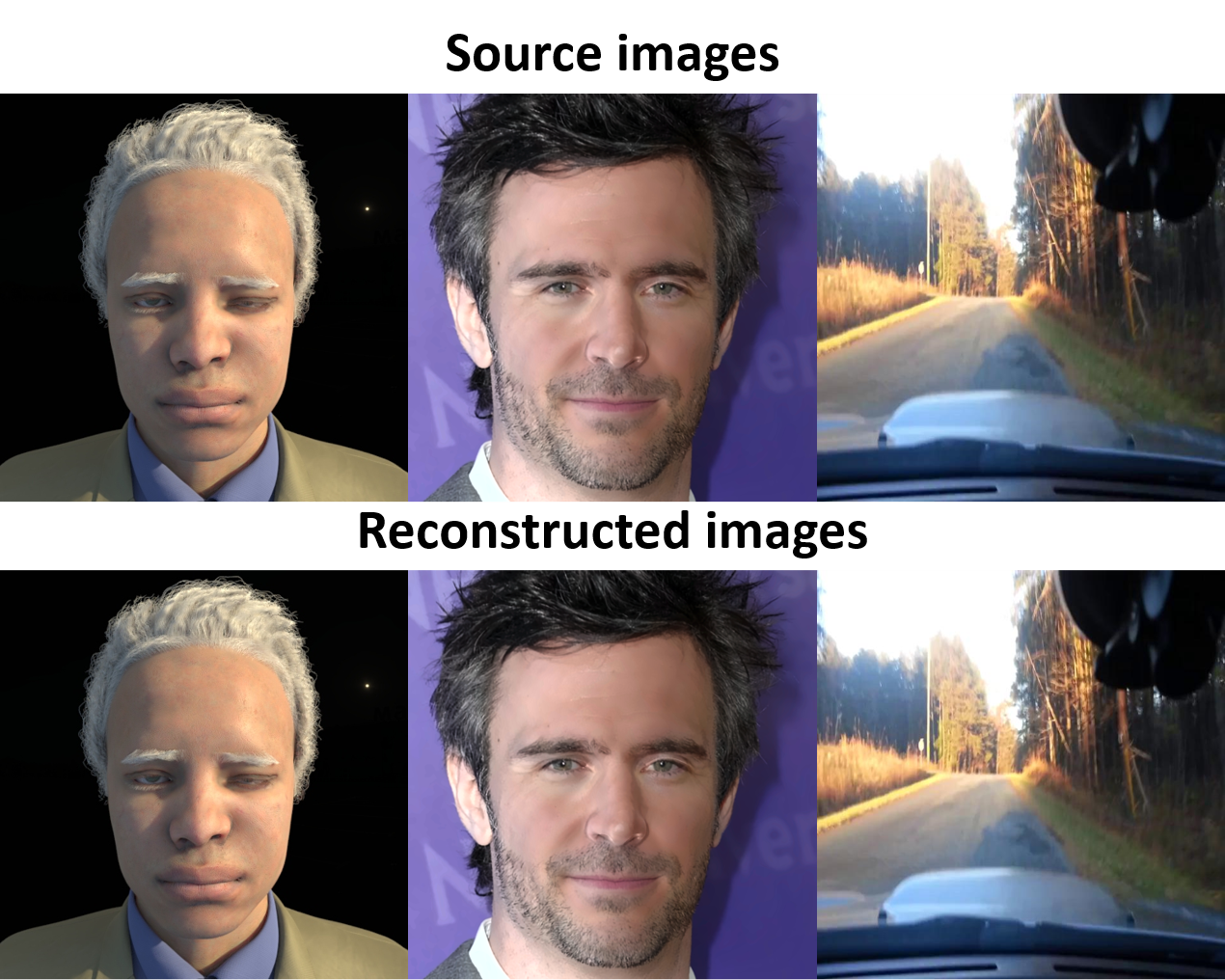}
    \caption{Reconstruction result of the reconstruction model build up two auto-encoder networks on the synthetic face, CelebA and Youtube 8M.} 
    \label{fig:reconstruction}
\end{figure}

\begin{figure}[hbtp]
    \centering
    \includegraphics[width=7 cm]{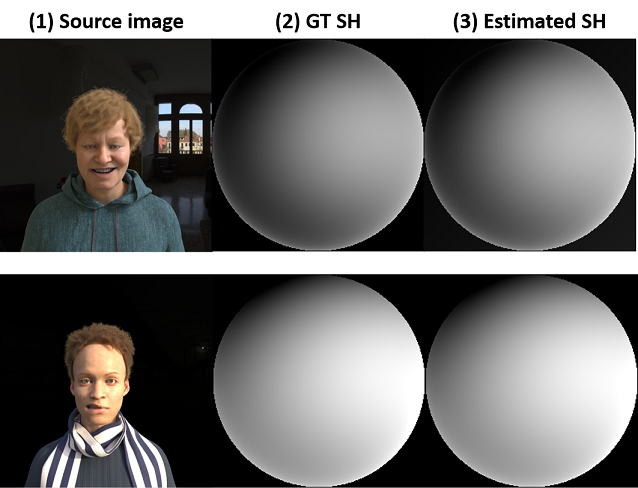}
    \caption{The estimated Spherical harmonic lighting of our method. The first column is the source images. The second column is the ground truth Spherical harmonic lighting. The third column is the estimated Spherical harmonic lighting of the source image. } 
    \label{fig:SHestimation}
\end{figure}

To understand the influence of the individual parts of the model, we remove them one at a time and evaluate the performance of the ablated model in Table \ref{tab:ablation} and Fig. \ref{fig:ablation}. Model (1) shows the performance of the full model with the reconstruction loss, SH loss and discriminator loss pre-trained on Youtube 8M and fine-tuning trained on the CelebA dataset. Model (2) does not use the reconstruction loss. Thus, it fails to light the image and some detail information is lost. The performance is lower than the full model. Model (3) does not use our proposed Spherical Harmonic Loss. In this case, the illumination embedding cannot be controlled in a semantically meaningful way. This also harms performance significantly. For further analysis, the performance SH loss, models (4-7) is shown. In rows (4-7), one of $\bm{A}_x$, $\bm{A}_y$, $\bm{A}_{xy}$ and $\bm{A}_{z}$ the SH loss is removed. Model (4) does not have a horizontal flip augmentation. Thus, the horizontal illumination can not be estimated and controlled and the performance decreases. Model (5) does not have a vertical flip, with a similar effect to Model (4). Model (6) shows the effect of $\bm{A}_{xy}$ is not significant, since only two channels are affected by $\bm{A}_{xy}$ in Eq. (6). Model (7) does not use the illumination inversion images, $\bm{A}_z$. The performance is influenced by over-exposure and under-exposure. Model (8) does not use Discriminator loss. We've found that the discriminator loss improves the visual quality of the relighted images and makes the relighted images sharp. Model (9) does not use the skip connection. Due to the limited space of embedding, some detail information lost. The accuracy drops significantly. As a result, we conclude that our full model can achieve a good balance between the accuracy and quality of the generated images.

\subsection{Reconstruction and Lighting Estimation}

\begin{figure}[hbtp]
    \centering
    \includegraphics[width=7 cm]{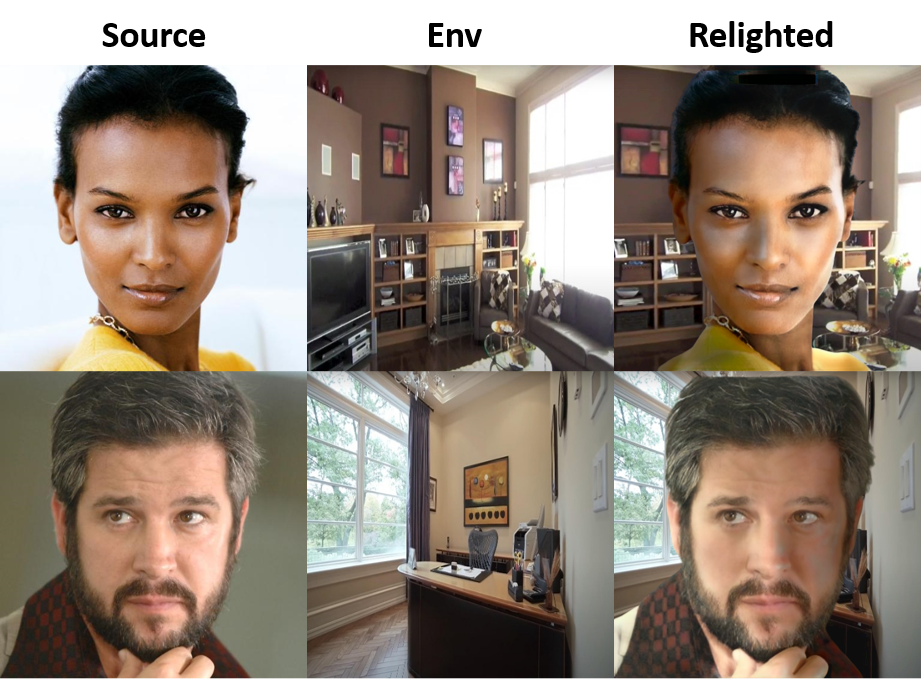}
    \caption{Visual lighting transformation results of the proposed method. The first column is the source images, the second column is the reference background images and the third column is the relighted images.} 
    \label{fig:background}
\end{figure}

In Fig. \ref{fig:reconstruction}, we show the reconstruction results based on the model shown in Fig. \ref{fig:R loss}. It is tested on synthetic face, CelebA and Youtube 8M, including generated face and real face. The reconstructed images contain fine details of the nose, eyes, mouth, even in the presence of extreme facial expression. The SH of the input image and estimated SH are shown in In Fig. \ref{fig:SHestimation}. Our method can estimate the SH light from a single background image. For further evaluating our proposed method, the images of Youtube 8M dataset are applied as the reference background images to relight face. The relighted images with the new background are shown in Fig. \ref{fig:background}. In our future work, we would test our proposed method on the asymmetric geometry and faces with strong shadows.

\subsection{Comparison with State-of-the-art Methods}

\begin{figure*}[hbtp]
    \centering
    \includegraphics[width=16 cm]{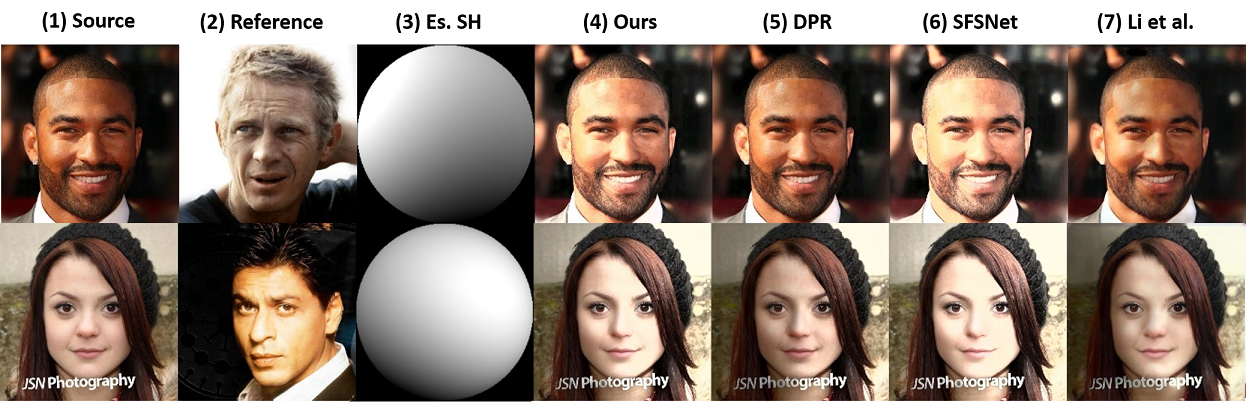}
    \caption{Visual lighting transformation results of the proposed method and state-of-the-art methods on CelebA dataset. The first three columns are the source image, the reference image and the estimated SH. Columns (4-7) shows the performance of the our proposed method, DPR \cite{zhou2019deep}, SFSNet \cite{sengupta2018sfsnet} and Li et al. \cite{li2018closed}, respectively.} 
    \label{fig:LightingTrans}
\end{figure*}

\begin{figure*}[hbtp]
    \centering
    \includegraphics[width=16 cm]{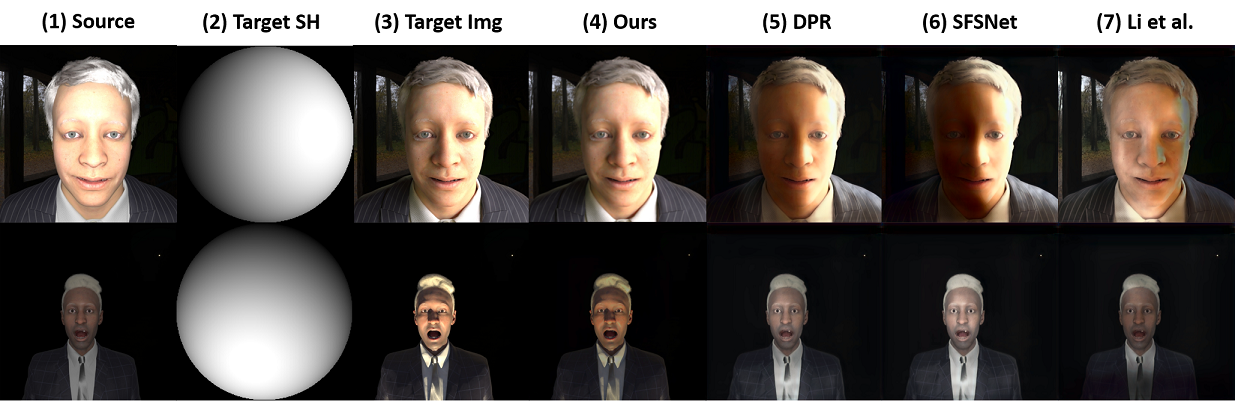}
    \caption{Visual lighting transformation results of the proposed method and state-of-the-art methods on synthetic face. The first three columns are the source image, the target SH and the target (ground truth) image. Columns (4-7) shows the performance of the our proposed method, DPR \cite{zhou2019deep}, SFSNet \cite{sengupta2018sfsnet} and Li et al. \cite{li2018closed}, respectively.} 
    \label{fig:LightingTrans2}
\end{figure*}

In this section, we compare our method with \cite{sengupta2018sfsnet, li2018closed, zhou2019deep} on the synthetic face and CelebA dataset. Since the illumination information in unknown in CelebA dataset, image-based relighting is used to evaluate our proposed method and baseline methods in Fig. \ref{fig:LightingTrans}. More specifically, target lighting used for relighting is extracted from a reference image. The proposed method, SFSNet \cite{sengupta2018sfsnet} and DPR \cite{zhou2019deep} can extract the lighting of the reference image with its own lighting estimation method. Li et al. \cite{li2018closed} is a state-of-the-art portrait style transfer method, which takes the source image and reference image as input and relights the source image as the illumination of the reference image. Since Li et al. and DPR are applied on the L channel of the input images, the input images are transferred from RGB image to Lab image. Our method and 
SFSNet are applied on the RGB channels of the input images. We notice that the edge of the relighted images of the proposed method is underexposed and some detail of face is lost. This is probably due to the gap between the training dataset and testing dataset. The results show that our proposed method can generate high-quality light estimations (1024 $\times$ 1024). Compared with DPR and SFSNet, our proposed method successfully overcomes the over-exposed images and the over-lighting problem of the nose and eyes. Li et al. do not generate images under the correct lighting. Since Li et al. uses reference images as input, the high-resolution images would improve the relighting performance while the low-resolution images can not estimate the reference lighting accurately. Since the illumination information is known in synthetic faces, relighting based on the SH is applied to evaluate our proposed method and baseline methods by comparing with the ground truth images in Fig. \ref{fig:LightingTrans2}. Our method can provide more face details and does not exhibit the over-lighting problem. In the second column, the skin colour of our relighted image is closer to the target image, since our proposed method input is a colour image, while the input to DPR is the L channel of the source image only. For further testing our proposed method, the RMSE-s between the target image and the relighted images of the different methods are calculated. The RMSE-s of ours (8.4 $\times 10^{-3}$) is only 79$\%$ of the RMSE-s of DPR (10.6$\times 10^{-3}$), SFSNet (10.8$\times 10^{-3}$) and Li et al. (11.3$\times 10^{-3}$). We also evaluate our proposed method on Multi-PIE shown in \href{https://github.com/yangliuav/Relighting-Based-Siamese-Auto-Encoder}{our Github page}, due to the limitation of pages.

\section{Conclusion}

We have proposed an automatic, unsupervised relighting algorithm trained on a large collection of unlabelled data. The relighting algorithm is build up by a Siamese Auto-encoder, where the source image information is split into content embedding and illumination embedding. Several auto-encoder networks are trained on the reconstruction, comparison and illumination estimation tasks. To provide target lighting, a Spherical Harmonic loss is first proposed, and four kinds of augmentation images are applied. We show that our training procedure, which combines reconstruction, Spherical Harmonic loss and adversarial losses, can estimate the illumination of the reference image and relight the source image. In addition, as our approach is trained on a large number of unlabelled data, it is less prone to exhibit common lighting artifacts and be applied on real as well as synthetic faces.

{\small
\bibliographystyle{ieee}
\bibliography{egbib}
}

\end{document}